\pdfoutput=1
\documentclass[11pt,a4paper]{article}
\usepackage{emnlp2021}
\usepackage{subcaption}
\usepackage{times}
\usepackage{graphicx}
\usepackage[T1]{fontenc}
\usepackage{longtable}
\usepackage{multicol}
\usepackage{multirow}
\usepackage{makecell}
\usepackage{placeins}
\usepackage{varioref}
\usepackage{algorithm}
\usepackage{microtype}
\usepackage{natbib}
\usepackage{latexsym}
\usepackage{amsmath}
\usepackage{float} 
\usepackage{amssymb}

\usepackage{amsthm}
\newtheorem*{remark}{Remark}
\newtheorem{theorem}{Theorem}
\theoremstyle{definition}
\newtheorem{definition}{Definition}[section]
\usepackage{algorithmicx}
\usepackage{algpseudocode}

\definecolor{lightRed}{RGB}{255, 128, 128}
\definecolor{red}{RGB}{255, 0, 0}
\definecolor{grey}{RGB}{179, 179, 179}
\definecolor{lightGreen}{RGB}{204, 255, 204}
\definecolor{green}{RGB}{0, 255, 2}

\algnewcommand\algorithmicinput{\textbf{INPUT:}}
\algnewcommand\INPUT{\item[\algorithmicinput]}

\algnewcommand\algorithmicoutput{\textbf{OUTPUT:}}
\algnewcommand\OUTPUT{\item[\algorithmicoutput]}
\usepackage{changes}

\algnewcommand\algorithmicbuild{\textbf{Build Negative Dataset:}}
\algnewcommand\NDS{\item[\algorithmicbuild]}
\algnewcommand\algorithminit{\textbf{Initialization:}}
\algnewcommand\Initialize{\item[\algorithminit]}

\algnewcommand\algorithStandardization{\textbf{Threshold:}}
\algnewcommand\Standardization{\item[\algorithStandardization]}

\algnewcommand\algorithoptim{\textbf{Optimization:}}
\algnewcommand\Optimization{\item[\algorithoptim]}
\usepackage{microtype}

\definecolor{ao}{rgb}{0.0, 0.5, 0.0}

\title{Improving Multimodal fusion via Mutual Dependency Maximisation} 

\author{\textbf{Pierre Colombo\textsuperscript{1,2}, Emile Chapuis\textsuperscript{1}}, \textbf{Matthieu Labeau\textsuperscript{\rm 1}, Chloe Clavel\textsuperscript{\rm 1}} \\
\textsuperscript{\rm 1}LTCI, Telecom Paris, Institut Polytechnique de Paris,\\ \textsuperscript{\rm 2}IBM GBS France,\\
\textsuperscript{\rm 1}firstname.lastname@telecom-paris.fr, \\
\textsuperscript{\rm 2}pierre.colombo@ibm.com,
}

\date{}
\begin{document}
\maketitle
\begin{abstract}
Multimodal sentiment analysis is a trending area of research, and the multimodal fusion is one of its most active topic. Acknowledging humans communicate through a variety of channels (i.e visual, acoustic, linguistic), multimodal systems aim at integrating different unimodal representations into a synthetic one. So far, a consequent effort has been made on developing complex architectures allowing the fusion of these modalities. However, such systems are mainly trained by minimising simple losses such as $L_1$ or cross-entropy. In this work, we investigate unexplored penalties and propose a set of new objectives that measure the dependency between modalities. We demonstrate that our new penalties lead to a consistent improvement (up to $4.3$ on accuracy) across a large variety of state-of-the-art models on two well-known sentiment analysis datasets: \texttt{CMU-MOSI} and \texttt{CMU-MOSEI}. Our method not only achieves a new SOTA on both datasets but also produces representations that are more robust to modality drops. Finally, a by-product of our methods includes a statistical network which can be used to interpret the high dimensional representations learnt by the model. 
\end{abstract}

\section{Introduction}\label{sec:introduction}
Humans employ three different modalities to communicate in a coordinated manner: the language modality with the use of words and sentences, the vision modality with gestures, poses and facial expressions and the acoustic modality through change in vocal tones. Multimodal representation learning has shown great progress in a large variety of tasks including emotion recognition, sentiment analysis \cite{multi_modal_sentiment}, speaker trait analysis \cite{pom} and fine-grained opinion mining \cite{multi_modal_opinion}. Learning from different modalities is an efficient way to improve performance on the target tasks \cite{xu2013survey}. Nevertheless, heterogeneities across modalities increase the difficulty of learning multimodal representations and raise specific challenges. \citet{tuto_multimodal} identifies fusion as one of the five core challenges in multimodal representation learning, the four other being: representation, modality alignment, translation and co-learning. Fusion aims at integrating the different unimodal representations into one common synthetic representation. Effective fusion is still an open problem: the best multimodal models in sentiment analysis \cite{multimodal_bert} improve over their unimodal counterparts, relying on text modality only, by less than 1.5\% on accuracy. Additionally, the fusion should not only improve accuracy but also make representations more robust to missing modalities. \\
Multimodal fusion can be divided into early and late fusion techniques: early fusion takes place at the feature level \cite{early_fusion}, while late fusion takes place at the decision or scoring level \cite{late_fusion}. Current research in multimodal sentiment analysis mainly focuses on developing new fusion mechanisms relying on deep architectures (\textit{e.g} \texttt{TFN} \cite{tfn}, \texttt{LFN} \cite{efficient_tfn}, \texttt{MARN} \cite{marne},  \texttt{MISA} \cite{misa}, \texttt{MCTN} \cite{MCTN}, \texttt{HFNN} \cite{hfnn}, \texttt{ICCN} \cite{ICCN}). Theses models are evaluated on several multimodal sentiment analysis benchmark such as \texttt{IEMOCAP} \cite{iemocap}, \texttt{MOSI} \cite{mosi}, \texttt{MOSEI} \cite{mosei} and \texttt{POM} \cite{pom_plus,pom}. Current state-of-the-art on these datasets uses architectures based on pre-trained transformers \cite{tsai2019multimodal,bert_like} such as MultiModal Bert (\texttt{MAGBERT}) or MultiModal XLNET (\texttt{MAGXLNET}) \cite{multimodal_bert}. 

The aforementioned architectures are trained by minimising either a $L_1$ loss or a Cross-Entropy loss between the predictions and the ground-truth labels. To the best of our knowledge, few efforts have been dedicated to exploring alternative losses. In this work, we propose a set of new objectives to perform and improve over existing fusion mechanisms. These improvements are inspired by the InfoMax principle \cite{infomax}, i.e. choosing the representation maximising the mutual information (MI) between two possibly overlapping views of the input. The MI quantifies the dependence of two random variables; contrarily to correlation, MI also captures non-linear dependencies between the considered variables. Different from previous work, which mainly focuses on comparing two modalities, our learning problem involves multiple modalities (\textit{e.g} text, audio, video). Our proposed method, which induces no architectural changes, relies on jointly optimising the target loss with an additional penalty term measuring the mutual dependency between different modalities.
\subsection{Our Contributions} We study new objectives to build more performant and robust multimodal representations through an enhanced fusion mechanism and evaluate them on multimodal sentiment analysis. Our method also allows us to explain the learnt high dimensional multimodal embeddings. The paper contributions can be summarised as follows:
\\\textbf{A set of novel objectives using multivariate dependency measures}. We introduce three new trainable surrogates to maximise the mutual dependencies between the three modalities (\textit{i.e} audio, language and video). We provide a general algorithm inspired by MINE \cite{mine}, which was developed in a bi-variate setting for estimating the MI. Our new method enriches MINE by extending the procedure to a multivariate setting that allows us to maximise different Mutual Dependency Measures: the Total Correlation \cite{total_correlation}, the f-Total Correlation and the Multivariate Wasserstein Dependency Measure \cite{bengio_wasserstein}. 
\\\textbf{Applications and numerical results}. We apply our new set of objectives to five different architectures relying on LSTM cells \cite{lstm} (\textit{e.g} \texttt{EF-LSTM}, \texttt{LFN}, \texttt{MFN}) or transformer layers (\textit{e.g} \texttt{MAGBERT}, \texttt{MAG-XLNET}). Our proposed method (1) brings a substantial improvement on two different multimodal sentiment analysis datasets (\textit{i.e} \texttt{MOSI} and \texttt{MOSEI},\autoref{ssec:results_loss}), (2) makes the encoder more robust to missing modalities (\textit{i.e} when predicting without language, audio or video the observed performance drop is smaller,~\autoref{ssec:robust}), (3) provides an explanation of the decision taken by the neural architecture (\autoref{ssec:qualitative_analysis}).

\section{Problem formulation \& related work}\label{sec:rw}
In this section, we formulate the problem of learning multi-modal representation (\autoref{ssec:represesentation}) and we review both existing measures of mutual dependency (see \autoref{ssec:mim}) and estimation methods (\autoref{ssec:estimation}). In the rest of the paper, we will focus on learning from three modalities (\textit{i.e} language, audio and video), however our approach can be generalised to any arbitrary number of modalities.

\subsection{Learning multimodal representations}\label{ssec:represesentation}
Plethora of neural architectures have been proposed to learn multimodal representations for sentiment classification.  Models often rely on a fusion mechanism (\textit{e.g} multi-layer perceptron \cite{late_fusion}, tensor factorisation \cite{efficient_tfn,FMT} or complex attention mechanisms \cite{mfn}) that is fed with modality-specific representations. The fusion problem boils down to learning a model $\mathcal{M}_f : \mathcal{X}_a \times \mathcal{X}_v \times \mathcal{X}_l \rightarrow \mathcal{R}^d$. $\mathcal{M}_f$ is fed with uni-modal representations of the inputs $X_{a,v,l} = (X_a,X_v,X_l)$ obtained through three embedding networks $f_a$, $f_v$ and $f_l$. $\mathcal{M}_f$ has to retain both modality-specific interactions (\textit{i.e} interactions that involve only one modality) and cross-view interactions (\textit{i.e} more complex, they span across both views). Overall, the learning of $\mathcal{M}_f$ involves both the minimisation of the downstream task loss and the maximisation of the mutual dependency between the different modalities.


\subsection{Mutual dependency maximisation}\label{ssec:mim}
\textbf{Mutual information as mutual dependency measure:}
the core ideas we rely on to better learn cross-view interactions are not new.  They consist of mutual information maximisation \cite{infomax}, and deep representation learning. Thus, one of the most natural choices is to use the MI that measures the dependence between two random variables, including high-order statistical dependencies \cite{true_mi}. Given two random variables $X$ and $Y$, the MI is defined by
\begin{equation}\label{eq:mi_def}
    I(X;Y) \triangleq \mathop{\mathbb{E}_{XY}} \left[ \log \frac{p_{XY}(x,y)}{p_X(x)p_Y(y)} \right],
\end{equation} where $p_{XY}$ is the joint probability density function (pdf) of the random variables $(X,Y)$, and $p_X$, $p_Y$ represent the marginal pdfs. MI can also be defined with a the KL divergence:
\begin{equation}\label{eq:mi_kl}
    I(X;Y) \triangleq \mathop{KL} \left[p_{XY}(x,y) || p_X(x)p_Y(y) \right].
\end{equation}
\textbf{Extension of mutual dependency to different metrics:} the KL divergence seems to be limited when used for estimating MI \cite{limitations_1}. A natural step is to replace the KL divergence in \autoref{eq:mi_kl} with different divergences such as the f-divergences or distances such as the Wasserstein distance. Hence, we introduce new mutual dependency measures (MDM): the f-Mutual Information \cite{mine}, denoted $I_f$ and the Wasserstein Measures \cite{bengio_wasserstein}, denoted $I_\mathcal{W}$. As previously, $p_{XY}$ denotes the joint pdf, and $p_X$, $p_Y$ denote the marginal pdfs. The new measures are defined as follows:
\begin{equation}\label{eq:wassser}
    I_f \triangleq \mathcal{D}_f(p_{XY}(x,y);p_X(x)p_Y(y)),
\end{equation} where  $\mathcal{D}_f$ denotes any $f$-divergences and 
\begin{equation}\label{eq:fdiver}
    I_\mathcal{W} \triangleq \mathcal{W}(p_{XY}(x,y);p_X(x)p_Y(y)),
\end{equation} where $\mathcal{W}$ denotes the Wasserstein distance \cite{peyre2019computational}.

\subsection{Estimating mutual dependency measures}\label{ssec:estimation}
\begin{figure}
    \centering
    \includegraphics[width=0.4\textwidth]{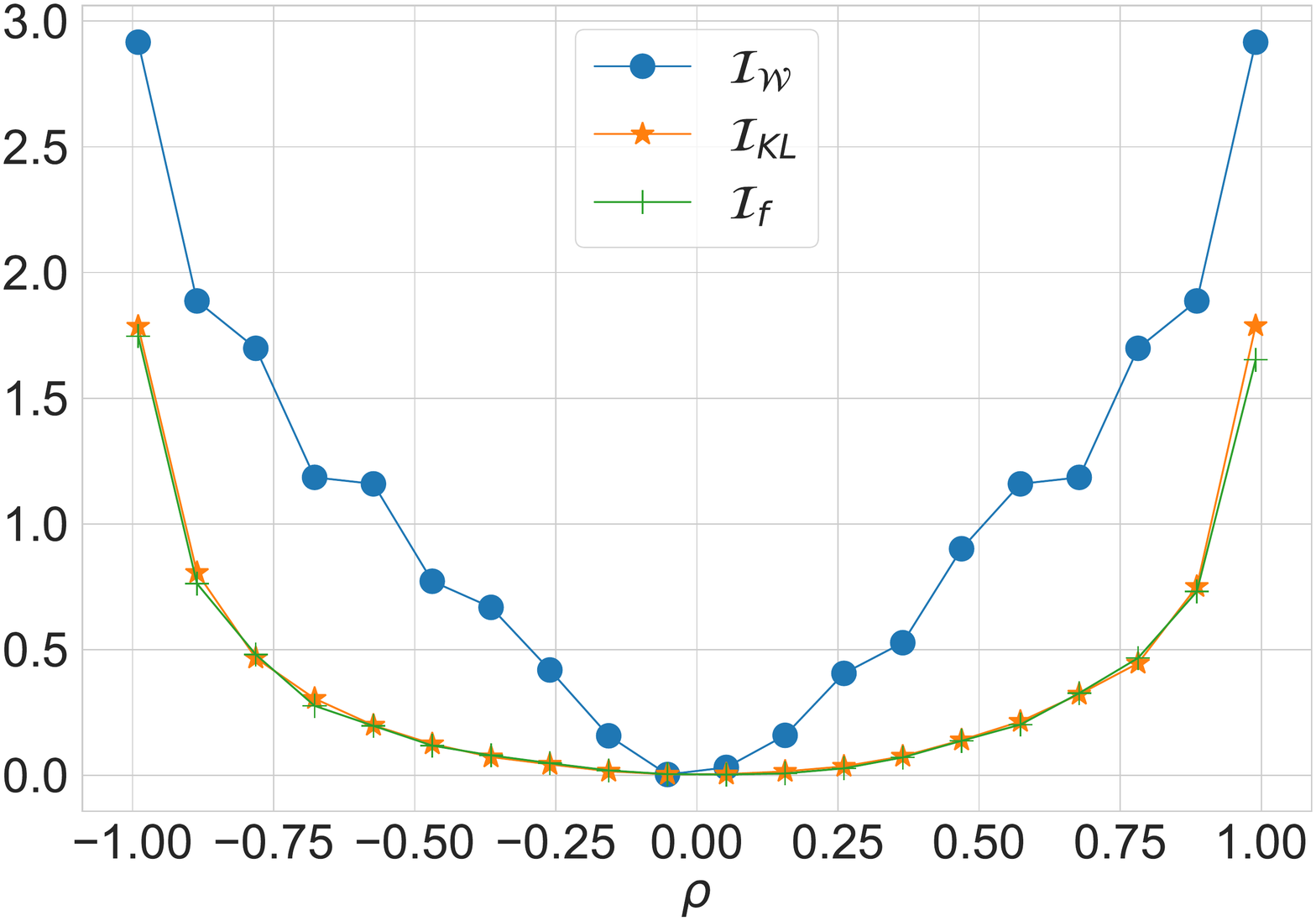}
    \caption{Estimation of different dependency measures for multivariate Gaussian random variables for different degree of correlation.}
    \label{fig:toy}
\end{figure}

The computation of MI and other mutual dependency measures can be difficult without knowing the marginal and joint probability distributions, thus it is popular to maximise lower bounds to obtain better representations of different modalities including image \cite{tian2019contrastive,hjelm2018learning}, audio \cite{dilpazir2016multivariate} and text \cite{kong2019mutual} data. Several estimators have been proposed: MINE \cite{mine} uses the Donsker-Varadhan representation \cite{donsker1985large} to derive a parametric lower bound holds, \citet{nguyen2017dual,nguyen2010estimating} uses variational characterisation of f-divergence and  a multi-sample version of the density ratio (also known as noise contrastive estimation \cite{nce,bengio_wasserstein}). These methods have mostly been developed and studied in a bi-variate setting.
\\\textbf{Illustration of neural dependency measures on a bivariate case}. In \autoref{fig:toy} we can see the aforementioned dependency measures (\textit{i.e} see \autoref{eq:mi_kl}, \autoref{eq:fdiver}, \autoref{eq:wassser}) when estimated with MINE \cite{mine} for multivariate Gaussian random variables, $X_a$ and $X_b$. The component wise correlation for the considered multivariate Gaussian is defined as follow: $corr(X_i, X_k) = \delta_{i,k} \rho$ , where $ \rho \in (-1, 1)$ and $\delta_{i,k}$ is Kronecker’s delta. We observe that the dependency measure based on Wasserstein distance is different from the one based on the divergences and thus will lead to different gradients. Although theoretical studies have been done on the use of different metrics for dependency estimations, it remains an open question to know which one is the best suited. In this work, we will provide an experimental response in a specific case. 

\section{Model and training objective}\label{sec:model}
In this section, we introduce our new set of losses to improve fusion. In \autoref{ssec:bivariate}, we first extend widely used bi-variate dependency measures to  multivariate dependencies \cite{james2017multivariate} measures (MDM). We then introduce variational bounds on the MDM, and in \autoref{ssec:theoritical}, we describe our method to minimise the proposed variational bounds.
\\\textbf{Notations} We consider $X_a, X_v, X_l$ as the multimodal data from the audio,video and language modality respectively with joint probability distribution $p_{X_{a}X_{v}X_{l}}$. We denote as $p_{X_j}$ the marginal distribution of $X_j$ with $j \in \{a,v,l\}$ corresponding to the $j$th modality.
\\\textbf{General loss} As previously mentioned, we rely on the InfoMax principle \cite{infomax} and aim at jointly maximising the MDM between the different modalities and minimising the task loss; hence, we are in a multi-task setting \cite{argyriou2007multi,ruder2017overview} and the objective of interest can be defined as: 

\begin{equation}\label{eq:total_loss}
    \mathcal{L} \triangleq \underbrace{\mathcal{L}_{down.}}_{\textrm{main task}} - \underbrace{\lambda \cdot \mathcal{L}_{MDM}}_{\textrm{mutual dependency term}}.
\end{equation}
$\mathcal{L}_{down.}$ represents a downstream  specific (target task) loss \textit{i.e} a binary cross-entropy or a $L_1$ loss, $\lambda$ is a meta-parameter and $\mathcal{L}_{MDM}$ is the multivariate dependencies measures (see \autoref{ssec:theoritical}). Minimisation of our newly defined objectives requires to derive lower bounds on the $\mathcal{L}_{MDM}$ terms, and then to obtain trainable surrogates.
\subsection{From bivariate to multivariate dependencies}\label{ssec:bivariate}
In our setting, we aim at maximising cross-view interactions involving three modalities, thus we need to generalise bivariate dependency measures to multivariate dependency measures. 
\begin{definition}[Multivariate Dependencies Measures]
\textit{Let $X_a, X_v, X_l$ be a set of random variables  with joint pdf $p_{X_{a}X_{v}X_{l}}$ and respective marginal pdf $p_{X_j}$ with $j \in \{a,v,l\}$. Then we defined {the multivariate mutual information} $\textbf{I}_{kl}$ which is also refered as total correlation \cite{total_correlation} or multi-information \cite{multiinformation}:}
\begin{align*}
    \textbf{I}_{kl} \triangleq \mathop{KL}(p_{X_aX_vX_l}(x_a,x_v,x_l)||\hspace{-0.5em}\prod\limits_{j\in\{a,v,l\}}\hspace{-0.5em}p_{X_j}(x_j)).
\end{align*}
\textit{Simarly for any f-divergence we define the {multivariate f-mutual information} $\textbf{I}_{f}$ as:}  \begin{align*}
    \textbf{I}_{f} \triangleq \mathcal{D}_f(p_{X_aX_vX_l}(x_a,x_v,x_l);\hspace{-0.5em}\prod\limits_{j\in\{a,v,l\}}\hspace{-0.5em}p_{X_j}(x_j)).
\end{align*} 
\textit{Finally, we also extend \autoref{eq:wassser} to obtain {the multivariate Wasserstein dependency} measure $\textbf{I}_\mathcal{W}$:}
\begin{align*}
    \textbf{I}_\mathcal{W} \triangleq \mathcal{W}(p_{X_aX_vX_l}(x_a,x_v,x_l);\hspace{-0.5em}\prod\limits_{j\in\{a,v,l\}}\hspace{-0.5em}p_{X_j}(x_j)).
\end{align*} \textit{where $\mathcal{W}$ denotes the Wasserstein distance.}
\end{definition}

\subsection{From theoretical bounds to trainable surrogates}\label{ssec:theoritical}
To train our neural architecture we need to estimate the previously defined multivariate dependency measures. We rely on neural estimators that are given in \autoref{the:estimator}.
\begin{theorem}\textbf{Multivariate Neural Dependency Measures}\label{the:estimator}
Let the family of functions $T(\theta) : \mathcal{X}_a \times \mathcal{X}_v  \times \mathcal{X}_l \rightarrow \mathbb{R}$  parametrized by a deep neural network with learnable parameters $\theta \in \Theta$. The multivariate mutual information measure $\textbf{I}_{kl}$ is defined as:
\begin{equation}\label{eq:kl}
        \textbf{I}_{kl} \triangleq\hspace{-0.1em} \underset{\theta}{\text{sup}} \quad\hspace{-0.9em} \mathbb{E}_{p_{X_aX_vX_l}} [T_\theta] - \log \big{[} \mathbb{E}_{\hspace{-1em}\prod\limits_{j\in\{a,v,l\}}\hspace{-1em}p_{X_j}} [e^{T_\theta}]\big].
\end{equation}
The neural multivariate f-mutual information measure $\textbf{I}_{f}$ is defined as follows:
\begin{equation}\label{eq:f-div}
\textbf{I}_{f} \triangleq  \underset{\theta}{\text{sup}}  \quad \mathbb{E}_{p_{X_aX_vX_l}} [T_\theta] - \mathbb{E}_{\hspace{-1em}\prod\limits_{j\in\{a,v,l\}}\hspace{-1em}p_{X_j}}[e^{T_\theta - 1}].
\end{equation}

The neural multivariate Wasserstein dependency measure $\textbf{I}_{\mathcal{W}}$ is defined as follows:
{\small
\begin{equation}\label{eq:wass}
\textbf{I}_{\mathcal{W}}\triangleq\hspace{-0.5em}\underset{\theta : T_\theta \in \mathbb{L}}{\text{sup}} \mathbb{E}_{p_{X_aX_vX_l}} [T_\theta] - \log \big[\mathbb{E}_{\hspace{-1em}\prod\limits_{j\in\{a,v,l\}}\hspace{-1em}p_{X_j}} [T_\theta]\big].
\end{equation}}
Where $\mathbb{L}$ is the set of all 1-Lipschitz functions from $\mathcal{R}^d \rightarrow \mathcal{R}$
\end{theorem}
\noindent\textbf{Sketch of proofs:} \autoref{eq:kl} is a direct application of the Donsker-Varadhan representation of the KL divergence (we assume that the integrability constraints are satisfied). \autoref{eq:f-div} comes from the work of \citet{nguyen2017dual}. \autoref{eq:wass} comes from the Kantorovich-Rubenstein: we refer the reader to \cite{villani2008optimal,peyre2019computational} for a rigorous and exhaustive treatment.
\\\textbf{Practical estimate of the variational bounds.} The empirical estimator that we derive from \autoref{the:estimator} can be used in practical way: the expectations in \autoref{eq:kl},~\autoref{eq:f-div} and \autoref{eq:wass} are estimated using empirical
samples from the joint distribution $p_{X_aX_vX_l}$. The empirical samples from $\prod\limits_{j\in\{a,v,l\}}p_{X_j}$ are obtained by shuffling the samples from the joint distribution in a batch.  We integrate this into minimising a multi-task objective \eqref{eq:total_loss} by using minus the estimator. We refer to the losses obtained with the penalty based on the estimators described in \autoref{eq:kl}, \autoref{eq:f-div} and \autoref{eq:wass} as $\mathcal{L}_{kl}$, $\mathcal{L}_f$ and $\mathcal{L}_\mathcal{W}$ respectively. Details on the practical minimisation of our variational bounds are provided in \autoref{alg:optimization}. 

\begin{remark}
In this work we choose to generalise MINE to compute multivariate dependencies. Comparing our proposed algorithm to other alternatives mentioned in \autoref{sec:rw} is left for future work. This choice is driven by two main reasons: (1) our framework allows the use of various types of contrast measures (\textit{e.g} Wasserstein distance,$f$-divergences); (2) the critic network $T_\theta$ can be used for interpretability purposes as shown in \autoref{ssec:qualitative_analysis}.
\end{remark}

{
\small
\begin{algorithm}[t]
 \begin{algorithmic}
\INPUT $\mathcal{D}_n=\{(x^{j}_a,x^{j}_v,x^{j}_l), \forall j \in [1,n]\}$ multimodal training dataset, $m$ batch size,  $\sigma_a,\sigma_v,\sigma_l : [1,m] \rightarrow [1,m]$ three permutations,  $\theta_c$ weights of the deep classifier, $\theta$ weights of the statistical network $T_\theta$.
\Initialize parameters $\theta$ and $\theta_c$
\NDS$$\bar{\mathcal{D}_n}=\{(x^{\sigma_a(j)}_a,x^{\sigma_v(j)}_v,x^{\sigma_l(j)}_l), \forall j \in [1,n]\}$$
\Optimization
\While{$(\theta,\theta_c)$ not converged}
\For{\texttt{ $i \in [1,Unroll]$}}
        \State Sample from $\mathcal{D}_n$, $\mathcal{B} \sim p_{X_aX_vX_l}$
                \State Sample from $\bar{\mathcal{D}_n}$, $\bar{\mathcal{B}} \sim \prod\limits_{j\in\{a,v,l\}}p_{X_j}$
        \State Update $\theta$ based on the empirical version of \autoref{eq:kl} or \autoref{eq:f-div} or \autoref{eq:wass}.
      \EndFor
\State Sample a batch $\mathcal{B}$ from $\mathcal{D}$
\State Update $\theta_c$ with $\mathcal{B}$ using \autoref{eq:total_loss}.
\EndWhile
\OUTPUT Classifiers weights ${\theta_c}$
\end{algorithmic}
\caption{Two-stage procedure to minimise multivariate dependency measures.}
\label{alg:optimization}
\end{algorithm}
}

\section{Experimental setting}\label{sec:experimental_setting}
In this section, we present our experimental settings including the neural architectures we compare, the datasets, the metrics and our methodology, which includes the hyper-parameter selection. 
\subsection{Datasets}\label{sec:mosi}
We empirically evaluate our methods on two english datasets: \texttt{CMU-MOSI} and \texttt{CMU-MOSEI}. Both datasets have been frequently used to assess model performance in human multimodal sentiment and emotion recognition.\\
\textbf{CMU-MOSI}: Multimodal Opinion Sentiment Intensity \cite{mosi} is a sentiment annotated dataset gathering $2,199$ short monologue video clips.\\
\textbf{CMU-MOSEI}: CMU-Multimodal Opinion Sentiment and Emotion Intensity \cite{mosei} is an emotion and sentiment annotated corpus consisting of $23,454$ movie review videos taken from  YouTube. Both \texttt{CMU-MOSI} and \texttt{CMU-MOSEI} are labelled by humans with a sentiment score in $[-3,3]$. 
For each dataset, three modalities are available; we follow prior work \cite{marne,tfn,multimodal_bert} and the features that have been obtained as follows\footnote{Data from \texttt{CMU-MOSI} and \texttt{CMU-MOSEI} can be obtained from \url{https://github.com/WasifurRahman/BERT_multimodal_transformer}}:
\\\textbf{Language}: Video transcripts are converted to word embeddings using either Glove \cite{pennington2014glove}, BERT or XLNET contextualised embeddings. For Glove, the embeddings are of dimension $300$, where for BERT and XLNET this dimension is $768$.
\\\textbf{Vision}: Vision features are extracted using Facet which results into facial action units corresponding to facial muscle movement. For  \texttt{CMU-MOSEI}, the video vectors are composed of $47$ units, and for \texttt{CMU-MOSI} they are composed of $35$.
\\ \textbf{Audio} : Audio features are extracted using COVAREP \cite{covarep}. This results into a vector of dimension $74$ which includes $12$ Mel-frequency cepstral coefficients (MFCCs), as well as pitch tracking and voiced/unvoiced segmenting features, peak slope parameters, maxima dispersion quotients and  glottal source parameters. \\ Video and audio are aligned on text-based following the convention introduced in  \cite{chen2017multimodal} and the forced alignment described in \cite{yuan2008speaker}.
\subsection{Evaluation metrics}
Multimodal Opinion Sentiment Intensity prediction is treated as a regression problem. Thus, we report both the Mean Absolute Error (MAE) and the correlation of model predictions with true labels. In the literature, the regression task is also turned into a binary classification task for polarity prediction. We follow standard practices \cite{multimodal_bert} and report the Accuracy\footnote{The regression outputs are turned into categorical values to obtain either $2$ or $7$ categories (see \cite{multimodal_bert,mfn,efficient_tfn})} ($Acc_7$ denotes accuracy on 7 classes and $Acc_2$ the binary accuracy) of our best performing models. 
\subsection{Neural architectures}
In our experiments, we choose to modify the loss function of the different models that have been introduced for multi-modal sentiment analysis on both \texttt{CMU-MOSI} and  \texttt{CMU-MOSEI}: Memory Fusion Network (\texttt{MFN} \cite{mfn}), Low-rank Multimodal Fusion (\texttt{LFN} \cite{efficient_tfn}) and two state-of-the-art transformers based models \cite{multimodal_bert} for fusion rely on BERT \cite{bert} (\texttt{MAG-BERT}) and XLNET \cite{xlnet} (\texttt{MAG-XLNT}). To assess the validity of the proposed losses, we also apply our method to a simple early fusion LSTM (\texttt{EF-LSTM}) as a baseline model.\\
\textbf{Model overview:} Aforementioned models can be seen as a multi-modal encoder $f_{\theta_e}$ providing a representation $Z_{avl}$ containing information and dependencies between modalities $X_l, X_a, X_v$ namely:
\begin{align*}
    f_{\theta_e}(X_a, X_v, X_l) &= Z_{avl}.
\end{align*}
As a final step, a linear transformation $A_{\theta_p}$ is applied to $Z_{avl}$ to perform the regression. 

\noindent \texttt{EF-LSTM}: is the most basic architecture used in the current multimodal analysis where each sequence view is encoded separately with LSTM channels. Then, a fusion function is applied to all representations.\\
\texttt{TFN}: computes a representation of each view, and then applies a fusion operator. Acoustic and visual views are first mean-pooled then encoded through a 2-layers perceptron. Linguistic features are computed with a LSTM channel. Here, the fusion function is a cross-modal product capturing unimodal, bimodal and trimodal interactions across modalities. \\
\texttt{MFN} enriches the previous \texttt{EF-LSTM} architecture with an attention module that computes a cross-view representation at each time step. They are then gathered and a final representation is computed by a gated multi-view memory~\cite{mfn}.\\
\texttt{MAG-BERT} and \texttt{MAG-XLNT} are based on pre-trained transformer architectures \cite{bert,xlnet} allowing inputs on each of the transformer units to be multimodal, thanks to a special gate inspired by \citet{wang2018words}. The $Z_{avl}$ is the $[CLS]$ representation provided by the last transformer head.  
For each architecture, we use the optimal architecture hyperparameters provided by the associated papers (see \autoref{sec:supplementary}).

\section{Numerical results}\label{sec:results}
We present and discuss here the results obtained using the experimental setting described in \autoref{sec:experimental_setting}. To better understand the impact of our new methods, we propose to investigate the following points: \\\textbf{Efficiency of the $\mathcal{L}_{MDM}$:} to gain understanding of the usefulness of our new objectives, we study the impact of adding the mutual dependency term on the basic multimodal neural model \texttt{EF-LSTM}.  
\\\textbf{Improving model performance and comparing multivariate dependency measures:}  the choice of the most suitable dependency measure for a given task is still an open problem (see \autoref{sec:model}). Thus, we compare the performance -- on both multimodal sentiment and emotion prediction tasks-- of the different dependency measures. The compared measures are combined with different models using various fusion mechanisms. 
\\\textbf{Improving the robustness to modality drop:} a desirable quality of multimodal representations is the robustness to a missing modality. We study how the maximisation of mutual dependency measures during training affects the robustness of the representation when a modality becomes missing.
\\\textbf{Towards explainable representations:} the statistical network $T_\theta$ allows us to compute a dependency measure between the three considered modalities. We carry out a qualitative analysis in order to investigate if a high dependency can be explained by complementariness across modalities. 

\subsection{Efficiency of the MDM penalty}\label{ssec:results_loss}
For a simple \texttt{EF-LSTM}, we study the improvement induced by addition of our MDM penalty. The results are presented in \autoref{tab:ef_lstm}, where a \texttt{EF-LSTM} trained with no  mutual dependency term is denoted with $\mathcal{L}_{\emptyset}$. On both studied datasets, we observe that the addition of a MDM penalty leads to stronger performances on all metrics. For both datasets, we observe that the best performing models are obtained by training with an additional mutual dependency measure term. Keeping in mind the example shown in~\autoref{fig:toy}, we can draw a first comparison between the different dependency measures. Although in a simple case $\mathcal{L}_{f}$  and $\mathcal{L}_{kl}$ estimate a similar quantity (see \autoref{fig:toy}), in more complex practical applications they do not achieve the same performance. Even though, the Donsker-Varadhan bound used for $\mathcal{L}_{kl}$ is stronger\footnote{For a fixed $T_\theta$ the right term in \autoref{eq:kl} is greater than \autoref{eq:f-div}} than the one used to estimate $\mathcal{L}_{f}$; for a simple model the stronger bound does not lead to better results. It is possible that most of the differences in performance observed come from the optimisation process during training\footnote{Similar conclusion have been drawn in the field of metric learning problem when comparing different estimates of the mutual information \cite{boudiaf2020unifying}.}.
\\\textbf{Takeaways:} On the simple case of \texttt{EF-LSTM} adding MDM penalty improves the performance on the downstream tasks.
\begin{table}[ht]
    \centering

 \resizebox{0.4\textwidth}{!}{\begin{tabular}{ |l||cccc| }
 \hline
 & $Acc^h_7$ & $Acc^h_2$ & $MAE^l$ & $Corr^h$\\
 \hline
\multicolumn{5}{c}{\texttt{CMU-MOSI}} \\
 \hline
 $\mathcal{L}_{\emptyset}$& 31.1 & 76.1 & 1.00 & 0.65\\
$\mathcal{L}_{kl}$& \underline{31.7} & \textbf{\underline{76.4}} & 1.00 & \textbf{\underline{0.66}}\\
$\mathcal{L}_{f}$  & \textbf{\underline{33.7}} & 76.2 & 1.02 & \textbf{\underline{0.66}}\\
$\mathcal{L}_{\mathcal{W}}$ & \underline{33.5}& \textbf{\underline{76.4}} & \textbf{\underline{0.98}} & \textbf{\underline{0.66}}\\
\hline
\multicolumn{5}{c}{\texttt{CMU-MOSEI}} \\
 \hline
$\mathcal{L}_{\emptyset}$ & 44.2 & 75.0 & 0.72 & 0.52\\
$\mathcal{L}_{kl}$ &  44.5 & \underline{75.6} & \underline{0.70} & \underline{0.53}\\
$\mathcal{L}_{f}$ & \textbf{4\underline{5.5}} & 75.2 & \underline{0.70} & 0.52\\
$\mathcal{L}_{\mathcal{W}}$ & \underline{45.3} & \textbf{\underline{75.9}} & \textbf{\underline{0.68}} & \textbf{\underline{0.54}}\\
    \hline
 \end{tabular}}
     \caption{Results on sentiment analysis on both \texttt{CMU-MOSI} and \texttt{CMU-MOSEI} for a \texttt{EF-LSTM}. $Acc_7$ denotes accuracy on 7 classes and $Acc_2$ the binary accuracy. $MAE$ denotes the Mean Absolute Error and $Corr$ is the Pearson correlation. $^h$ means higher is better and $^l$ means lower is better. The choice of the evaluation metrics follows standard practices \cite{multimodal_bert}. Underline results demonstrate significant improvement (p-value belows 0.05) against the baseline when performing the Wilcoxon Mann Whitney test \cite{wilcoxon1992individual} on 10 runs using different seeds.}
    \label{tab:ef_lstm}
 \end{table}
 
\subsection{Improving models and comparing multivariate dependency measures}
In this experiment, we apply the different penalties to more advanced architectures, using various fusion mechanisms. 
\\\textbf{General analysis}. \autoref{tab:performances} shows the performance of various neural architectures trained with and without MDM penalty. Results are coherent with the previous experiment: we observe that jointly maximising a mutual dependency measure leads to better results on the downstream task: for example, a \texttt{MFN} on \texttt{CMU-MOSI} trained with $\mathcal{L}_{\mathcal{W}}$  outperforms by $4.6$ points on $Acc^h_7$ the model trained without the mutual dependency term. On \texttt{CMU-MOSEI} we also obtain subsequent improvements while training with MMD. On \texttt{CMU-MOSI} the \texttt{TFN} also strongly benefits from the  mutual dependency term with an absolute improvement of $3.7\%$ (on $Acc^h_7$) with $\mathcal{L}_{\mathcal{W}}$ compared to $\mathcal{L}_{\emptyset}$. 
\autoref{tab:performances} shows that our methods not only perform well on recurrent architectures but also on pretrained Transformer-based models, that achieve higher results due to a superior capacity to model contextual dependencies (see \cite{multimodal_bert}). 
\\\textbf{Improving state-of-the-art models}. \texttt{MAGBERT} and \texttt{MAGXLNET} are state-of-the art models on both \texttt{CMU-MOSI} and \texttt{CMU-MOSEI}. From \autoref{tab:performances}, we observe that our methods can improve the performance of both models. It is worth noting that, in both cases, $\mathcal{L}_{\mathcal{W}}$ combined with pre-trained transformers achieves good results. This performance gain suggests that our method is able to capture dependencies that are not learnt during either pretraining of the language model (\textit{i.e} BERT or XLNET) or by the Multimodal Adaptation Gate used to perform the fusion.
\\\textbf{Comparing dependency measures}. \autoref{tab:performances} shows that there is no dependency measure that achieves the best results in all cases. This result tends to confirm that the optimisation process during training plays an important role (see hypothesis in \autoref{ssec:results_loss}). However, we can observe that optimising the multivariate Wasserstein dependency measure is usually a good choice, since it achieves state of the art results in many configurations. It is worth noting that several pieces of research point the limitations of mutual information estimators \cite{limitations_1,limitations_2}.
\\\textbf{Takeaways:} The addition of MMD not only benefits simple models (\textit{e.g} \texttt{EF-LSTM}) but also improves performance when combined with both complex fusion mechanisms and pretrained models. For practical applications, the Wasserstein distance is a good choice of contrast function.
\begin{table}

    \centering

 \resizebox{0.47\textwidth}{!}{\begin{tabular}{ |l||cccc||cccc| }
 \hline
  &\multicolumn{4}{c}{\texttt{CMU-MOSI}} & \multicolumn{4}{c}{\texttt{CMU-MOSEI}} \\
 \hline
 & $Acc^h_7$ & $Acc^h_2$ & $MAE^l$ & $Corr^h$  & $Acc^h_7$ & $Acc^h_2$ & $MAE^l$ & $Corr^h$\\
 \hline
\multicolumn{9}{c}{\quad\quad\texttt{MFN}} \\
 \hline
   $\mathcal{L}_{\emptyset}$ & 31.3 & 76.6 & 1.01 & 0.62& 44.4 & 74.7 & 0.72 & 0.53 \\
   $\mathcal{L}_{kl}$& \underline{32.5} & 76.7 & \textbf{\underline{0.96}} & \textbf{0.65}& 44.2 & 74.7 & 0.72 & \textbf{\underline{0.57}} \\
    $\mathcal{L}_{f}$  & \underline{35.7} & \underline{77.4} & \textbf{\underline{0.96}} & \textbf{0.65} & \underline{46.1} & \textbf{75.4} & \textbf{\underline{0.69}} & \underline{0.56} \\
    $\mathcal{L}_{\mathcal{W}}$ & \textbf{\underline{35.9}} & \textbf{\underline{77.6}} & \textbf{\underline{0.96}} & \textbf{0.65} & \textbf{\underline{46.2}} & 75.1 & \textbf{\underline{0.69}} & \underline{0.56} \\
     \hline
\multicolumn{9}{c}{\quad\quad\texttt{LFN}} \\
 \hline
    $\mathcal{L}_{\emptyset}$ & 31.9 & 76.9 & 1.00 & 0.63   
    &45.2 & 74.2 & 0.70 & 0.54 \\
    
    $\mathcal{L}_{kl}$& \underline{32.6} & \textbf{\underline{77.7}} & 0.97 & 0.63 
    
    & \underline{46.1} & 75.3 & 0.68 & \textbf{\underline{0.57}}  \\
    
    $\mathcal{L}_{f}$  &  \textbf{\underline{35.6}} & 77.1 & 0.97 & 0.63  
    & 45.8 & \textbf{\underline{75.4}}& 0.69 & \textbf{\underline{0.57}} \\
    
   $\mathcal{L}_{\mathcal{W}}$ & \textbf{\underline{35.6}} & \textbf{\underline{77.7}} & \textbf{\underline{0.96}} & \textbf{\underline{0.67}}  
   & \textbf{\underline{46.2}} & \textbf{\underline{75.4}} & \textbf{\underline{0.67}} & \textbf{\underline{0.57}} \\
         \hline
\multicolumn{9}{c}{\quad\quad\texttt{MAGBERT}} \\
 \hline
    $\mathcal{L}_{\emptyset}$ & 40.2 & 84.7 & 0.79 & 0.80 & 46.8 & 84.9 & \textbf{0.59} & 0.77  \\
    
   $\mathcal{L}_{kl}$& \textbf{\underline{42.0}} & \textbf{\underline{85.6}} & \textbf{\underline{0.76}} & \textbf{0.82}& 47.1 & 85.4 & \textbf{0.59} & \textbf{\underline{0.79}} \\
   
   $\mathcal{L}_{f}$  & \underline{41.7} & \textbf{\underline{\underline{85.6}}} & 0.78 & \textbf{0.82}&  46.9 & \textbf{85.6} & \textbf{0.59} & \textbf{\underline{0.79}} \\
   
    $\mathcal{L}_{\mathcal{W}}$& \underline{41.8} & 85.3 & \textbf{\underline{0.76}} & \textbf{0.82}& \textbf{\underline{47.8}} & 85.5 & \textbf{0.59} & \textbf{\underline{0.79}}  \\
             \hline
\multicolumn{9}{c}{\quad\quad\texttt{MAGXLNET}} \\
 \hline
          $\mathcal{L}_{\emptyset}$ & 43.0 & 86.2 & 0.76 & \textbf{0.82}
         
          & 46.7 & 84.4 & \textbf{0.59} & 0.79 \\
    $\mathcal{L}_{kl}$& \textbf{\underline{44.5}} & 86.1 & \textbf{\underline{0.74}} & \textbf{0.82}
    
    &\underline{47.5} & \underline{85.4} & \textbf{0.59} & 0.81  \\
    
    $\mathcal{L}_{f}$  & \underline{43.9} & 86.6 & \textbf{\underline{0.74}} & \textbf{0.82}& 
    
    47.4 & 85.0 & \textbf{0.59} & 0.81 \\
    $\mathcal{L}_{\mathcal{W}}$ & \underline{44.4} & \textbf{\underline{86.9}} & \textbf{\underline{0.74}} &\textbf{0.82} 
    & \textbf{\underline{47.9}}& \textbf{\underline{85.8}} & \textbf{0.59} & \textbf{\underline{0.82}}\\
    \hline
 \end{tabular}}
     \caption{Results on sentiment and emotion prediction on both \texttt{CMU-MOSI} and  \texttt{CMU-MOSEI} dataset for the different neural architectures presented in \autoref{sec:experimental_setting} relying on various fusion mechanisms. }
    \label{tab:performances}

\end{table}


\begin{figure}[htb!]
\centering
\includegraphics[width=0.47\textwidth]{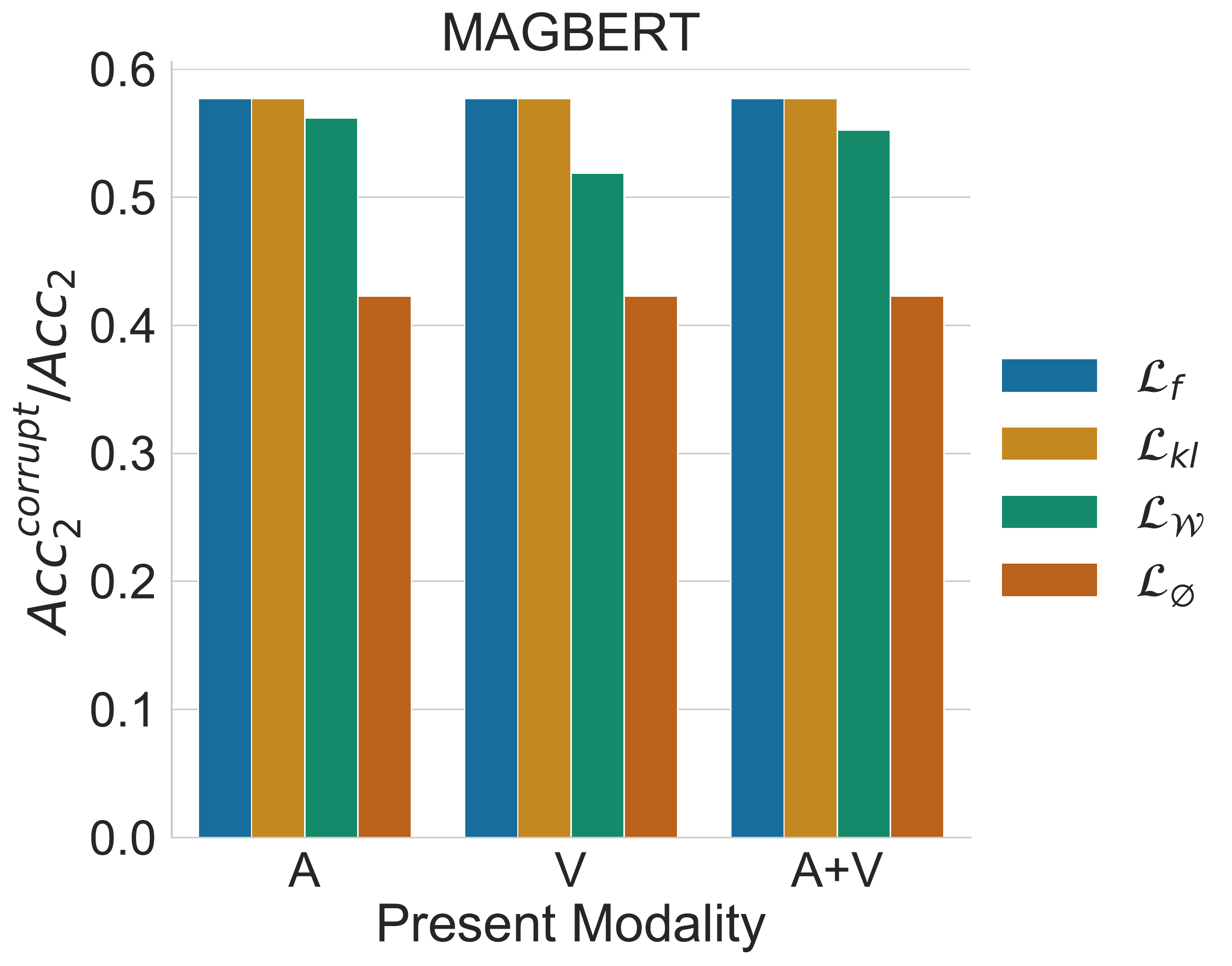}
\caption{Study of the robustness of the representations against drop of the linguistic modality. Studied model is \texttt{MAGBERT} on \texttt{CMU-MOSI}. The ratio between the accuracy achieved with a corrupted linguistic modality $Acc_2^{corrupt}$ and the accuracy $Acc_2$ without any corruption is reported on $y$-axis. The preserved modalities during inference are reported on $x$-axis. $A$, $V$ respectively stands for acoustic and visual modality.}

\label{fig:modality_drop_bert_summary}
\vspace{-.4cm}
\end{figure}
 \begin{table*}
    \centering
    \resizebox{\textwidth}{!}{
    \begin{tabular}{l l|c}\hline
       Spoken Transcripts & Acoustic and visual behaviour & $T_\theta$  \\\hline
      \colorbox{lightGreen}{um the story was all right}  & \colorbox{lightRed}{low energy monotonous voice} + \colorbox{grey}{headshake} & L\\
           \colorbox{green}{i mean its a Nicholas Sparks book it must be good}  & \colorbox{lightRed}{disappointed tone} + \colorbox{grey}{neutral facial expression}& L \\\hline
                \colorbox{green}{the action is fucking awesome} & \colorbox{lightGreen}{head nod} + \colorbox{green}{excited voice}  & H \\
                    \makecell[l]{
                    \colorbox{lightGreen}{it was cute you know the actors  did a great job bringing the smurfs to} \\ \colorbox{lightGreen}{life such as joe george  lopez neil patrick harris katy perry and a fourth}} 
                    & \colorbox{green}{multiple smiles} & H\\ \hline
    \end{tabular}}
    \caption{Examples from the \texttt{CMU-MOSI} dataset using \texttt{MAGBERT}. The last column is computed using the statistical network $T_\theta$. $L$ stands for low values and $H$ stands for high values. Green, grey, red highlight positive, neutral and negative expression/behaviours respectively}
    \label{tab:my_label}\vspace{-.3cm}
\end{table*}
\subsection{Improved robustness to modality drop}\label{ssec:robust}
Although fusion with visual and acoustic modalities provided a performance improvement \cite{wang2018words}, the performance of Multimodal systems on sentiment prediction tasks is mainly carried by the linguistic modality \cite{mfn,tfn}. Thus it is interesting to study how a multimodal system behaves when the text modality is missing because it gives insights on the robustness of the representation.
\\\textbf{Experiment description}. In this experiment, we focus on the \texttt{MAGBERT} and \texttt{MAGXLNET} since they are the best performing models.\footnote{Because of space constraints results corresponding to \texttt{MAGXLNET} are reported in \autoref{sec:supplementary}.} As before, the considered models are trained using the losses described in \autoref{sec:model} and all modalities are kept during training time. During inference, we either keep only one modality (Audio or Video) or both. Text modality is always dropped.
\\\textbf{Results}. Results of the experiments conducted on \texttt{CMU-MOSI} are shown in \autoref{fig:modality_drop_bert_summary}, giving values for the ratio $Acc_2^{corrupt}/Acc_2$ where $Acc_2^{corrupt}$ is the binary accuracy in the corrupted configuration and $Acc_2$ the accuracy obtained when all modalities are considered. We observe that models trained with an MDM penalty (either $\mathcal{L}_{kl}$, $\mathcal{L}_{f}$ or $\mathcal{L}_\mathcal{W}$) resist better to missing modalities than those trained with $\mathcal{L}_\emptyset$. For example, when trained with $\mathcal{L}_{kl}$ or $\mathcal{L}_{f}$, the drop in performance is limited to $\approx 25\%$ in any setting. Interestingly, for \texttt{MAGBERT} $\mathcal{L}_\mathcal{W}$ and $\mathcal{L}_{KL}$ achieve comparable results; $\mathcal{L}_{KL}$ is more resistant to dropping the language modality, and thus, could be preferred in practical applications.
\\\textbf{Takeaway:} Maximising the MMD allows an information transfer between modalities.





\subsection{Towards explainable representations}\label{ssec:qualitative_analysis}
In this section, we propose a qualitative experiment allowing us to interpret the predictions made by the deep neural classifier. During training, $T_\theta$ estimates the mutual dependency measure, using the surrogates introduced in \autoref{the:estimator}. However, the inference process only involves the classifier, and $T_\theta$ is unused. \autoref{eq:kl}, \autoref{eq:f-div},~\autoref{eq:wass} show that $T_\theta$ is trained to discriminate between valid representations (coming from the joint distribution) and corrupted representations (coming from the product of the marginals). Thus, $T_\theta$ can be used, at inference time, to measure the mutual dependency of the representations used by the neural model. In \autoref{tab:my_label} we report examples of low and high discrepancy measures for \texttt{MAGBERT} on \texttt{CMU-MOSI}. We can observe that high values correspond to video clips where audio, text and video are complementary (\textit{e.g} use of head node \cite{head_nod}) and low values correspond to the case where there exists contradictions across several modalities. Results on \texttt{MAGXLET} can be found in \autoref{ssec:supplementary_qualitative_examples}.
\\\textbf{Takeaways:} $T_\theta$ used to estimate the MDM provides a mean to interpret representations learnt by the encoder.

\section{Conclusions}\label{sec:conclusion}
In this paper, we introduced three new losses based on MDM. Through extensive set of experiments on \texttt{CMU-MOSI} and \texttt{CMU-MOSEI}, we have shown that SOTA architectures can benefit from these innovations with little modifications. A by-product of our method involves a statistical network that is a useful tool to explain the learnt high dimensional multi-modal representations. This work paves the way for using and developing new alternative methods to improve the learning (\textit{e.g} new estimator of mutual information \cite{DBLP:journals/corr/abs-2105-02685}, Wasserstein Barycenters \cite{DBLP:journals/corr/abs-2108-12463}, Data Depths \cite{staerman2021affine}, Extreme Value Theory \cite{DBLP:conf/nips/JalalzaiCCGVVS20}). A future line of research involves using this methods for emotion \cite{DBLP:conf/naacl/ColomboWMKK19,DBLP:conf/wassa/WitonCMK18} and dialog act \cite{DBLP:journals/corr/abs-2108-12465,DBLP:journals/corr/abs-2009-11152,DBLP:conf/emnlp/ChapuisCMLC20} classification with pre-trained model tailored for spoken language \cite{DBLP:conf/emnlp/DinkarCLC20}.

\section{Acknowledgments}
The research carried out in this paper has received funding from IBM, the French National Research Agency’s grant ANR-17-MAOI and the DSAIDIS chair at Telecom-Paris. This work was also granted access to the HPC resources of IDRIS under the allocation 2021-AP010611665 as well as under the project 2021-101838 made by GENCI. 
\bibliography{acl2020}
\bibliographystyle{acl_natbib}
\newpage
\clearpage
\section{Supplementary}\label{sec:supplementary}
\subsection{Training details}
In this section, we both present a comprehensive illustration of the \autoref{alg:optimization} and state the details of experimental hyperparameters selection as well as and the architectures used for the statistic network $T_\theta$.

\subsubsection{Illustration of \autoref{alg:optimization}} 
\begin{figure}[!ht]
    \centering
    \includegraphics[width=0.5\textwidth]{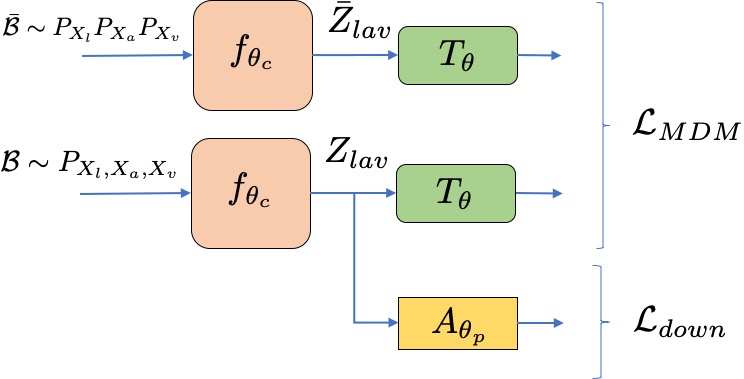}
    \caption{Illustration of the method describes in \autoref{alg:optimization} for the different estimators derived from \autoref{the:estimator}. $\mathcal{B}$ and $\bar{\mathcal{B}}$ stands for the batch of data sample from the joint probability distribution and the product of the marginal distribution respectively.  $Z_{avl}$ denotes the fusion representation of linguistic, acoustic and visual (resp. $l$, $a$ and $v$) modalities provided by a multimodal architecture $f_{\theta_e}$ for the batch $\mathcal{B}$ . $\overline{Z}_{lav}$ denotes the same quantity as described before for the batch $\bar{\mathcal{B}}$. $A_{\theta_p}$ denotes the linear projection before classification or regression.}
    \label{fig:illustration}
\end{figure}

\autoref{fig:illustration} describes the \autoref{alg:optimization}. As can be seen in the figure, to compute the mutual dependency measure the statistic network $T_\theta$ takes the two embeddings of the different batch $\mathcal{B}$ and $\bar{\mathcal{B}}$.

\subsubsection{Hyperparameters selection}
 We use dropout \cite{dropout} and optimise the global loss \autoref{eq:total_loss} by gradient descent using AdamW \cite{adamw,adam} optimiser. The best learning rate is found in the grid $\{0.002,0.001,0.0005,0.0001\}$. The best model is selected using the lowest MAE on the validation set. We $Unroll$ to 10. 
\subsubsection{Architectures of $T_\theta$}
Across the different experiment we use a statistic network with an architecture as describes in \autoref{tab:stat_net}. We follow \cite{mine} and use LeakyRELU \cite{relu,xu2015empirical} as activation function. 
\begin{table}[!ht]
    \centering
 \resizebox{0.46\textwidth}{!}{\begin{tabular}{ l|c|c }
 \hline
 \multicolumn{3}{c}{Statistic Network} \\
 \hline
 Layer & Number of outputs & Activation function\\
 \hline
 $[Z_{lav}, \overline{Z}_{lav}]$ & $d_{in}, d_{in}$& -\\
 Dense layer & $d_{in}/2$ & LeakyReLU\\
 Dropout & 0.4  &  -\\
  Dense layer & $d_{in}$ & LeakyReLU\\
 Dropout & 0.4  &  -\\
  Dense layer & $d_{in}$ & LeakyReLU\\
 Dropout & 0.4  &  -\\
 Dense layer& $d_{in}/4$ &LeakyReLU \\
 Dropout & 0.4 & - \\
 Dense layer & $d_{in}/4$ & LeakyReLU\\
 Dropout & 0.4 &  -\\
 Dense layer & 1 & Sigmoïd \\
 \hline
 \end{tabular}}
     \caption{Statistics network description. $d_{in}$ denotes the dimension of $Z_{avl}$.}
    \label{tab:stat_net}
 \end{table}

\subsection{Additional experiments for robustness to modality drop}
\begin{figure}[ht!]
\centering
\includegraphics[width=0.4\textwidth]{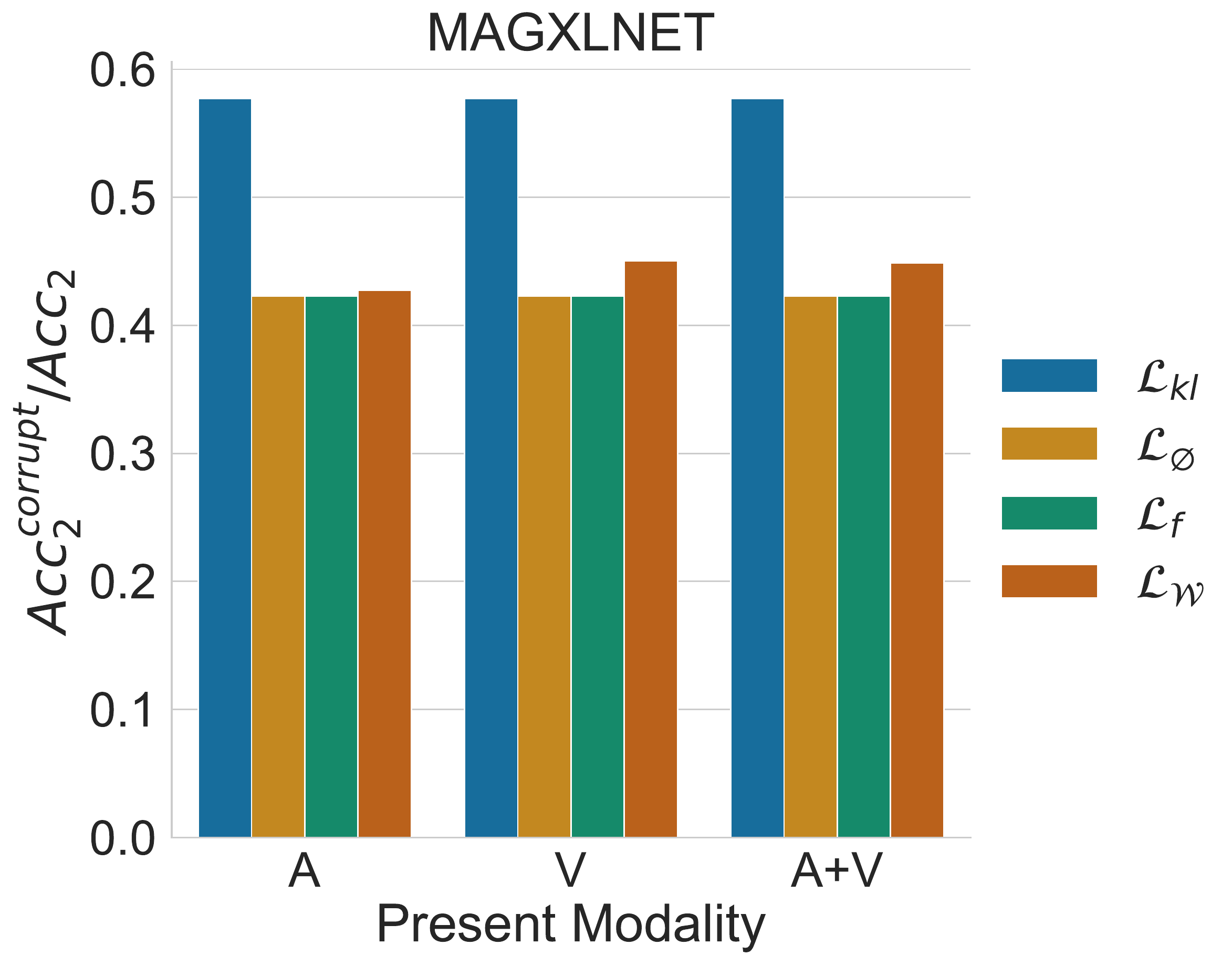}
\caption{Study of the robustness of the representations against a drop of the linguistic modality. Studied model is \texttt{MAGXLNET} on \texttt{CMU-MOSI}. The ratio between the accuracy achieved with a corrupted linguistic modality $Acc_2^{corrupt}$ and the accuracy $Acc_2$ without any corruption is reported on $y$-axis. The preserved modalities during inference are reported on $x$-axis. $A$, $V$ respectively stands for the acoustic and visual modality.}\label{fig:magxlnet}
\end{figure}

\autoref{fig:magxlnet} shows the results of the robustness text on \texttt{MAGXLNET}. Similarly to \autoref{fig:modality_drop_bert_summary} we observe more robust representation to modality drop when jointly maximising the $\mathcal{L}_\mathcal{W}$ and $\mathcal{L}_{kl}$ with the target loss. \autoref{fig:magxlnet} shows no improvement when training with $\mathcal{L}_f$. This can also be linked to \autoref{tab:performances} which similarly shows no improvement in this very specific configuration. 

\subsection{Additional qualitative examples}\label{ssec:supplementary_qualitative_examples}
\autoref{tab:qualitative_xlnet} illustrates the use of $T_\theta$ to explain the representations learnt by the model. Similarly to \autoref{tab:stat_net} we observe that high values correspond to complementarity across modalities and low values are related to contradictoriness across modalities.
 \begin{table*}[ht]
    \resizebox{0.9\textwidth}{!}{
    \begin{tabular}{l l|c}\hline
      Spoken Transcripts & Acoustic and visual behaviour & $T_\theta$  \\\hline
      \colorbox{lightRed}{but the m the script is corny}  & \colorbox{lightGreen}{high energy voice} + \colorbox{grey}{headshake} +  \colorbox{green}{(many) smiles}  & L\\
                          \makecell[l]{
                    \colorbox{red}{as for gi joe was it was just like laughing } \\ \colorbox{red}{its the the plot the the acting is terrible}} 
                     & \colorbox{lightGreen}{high enery voice} + 
                     \colorbox{green}{laughts} +
                     \colorbox{green}{smiles} & L \\
                      \colorbox{green}{but i think this one did beat scream 2 now}  & \colorbox{grey}{headshake} + \colorbox{red}{long sigh}& L \\
                                            \colorbox{green}{the xxx sequence is really well done}  & \colorbox{grey}{static head} + \colorbox{red}{low energy monotonous voice}& L \\
          \hline

                    \colorbox{green}{you know of course i was waithing for the princess and the frog} & \colorbox{green}{smiles} + \colorbox{green}{high energy voice} + + \colorbox{green}{high pitch} & H\\ 
                    \colorbox{green}{dennis quaid i think had a lot of fun} & \colorbox{green}{smiles} + \colorbox{green}{high energy voice} & H\\ 
                                    \colorbox{red}{it was very very very boring} & \colorbox{red}{low energy voice} + \colorbox{red}{frown eyebrows}  & H \\
                    \colorbox{red}{i do not wanna see any more of this} & \colorbox{red}{angry voice} + \colorbox{red}{angry facial expression} & H\\ \hline
    \end{tabular}}
    \caption{Examples from the \texttt{CMU-MOSI} dataset using \texttt{MAGXLNET} trained with $\mathcal{L}_\mathcal{W}$. The last column is computed using the statistic network $T_\theta$. $L$ stands for low values and $H$ stands for high values. Green, grey, red highlight positive, neutral and negative expression/behaviours respectively.}
    \label{tab:qualitative_xlnet}
\end{table*}

\end{document}